\documentclass[conference]{IEEEtran}

\IEEEoverridecommandlockouts
\usepackage{cite}
\usepackage{amsmath,amssymb,amsfonts}
\usepackage{algorithmic}
\usepackage{graphicx}
\usepackage{textcomp}
\usepackage{xcolor}
\usepackage{amsmath}
\usepackage[style=base]{caption}
\usepackage[utf8]{inputenc}
\usepackage{gensymb}
\usepackage[font=small,labelfont=bf]{caption} % Required for specifying captions to tables and figures
\definecolor{cb_red}{rgb}{0.89,0.1,0.11}
\definecolor{brown}{rgb}{0.59, 0.29, 0.0}

\def\BibTeX{{\rm B\kern-.05em{\sc i\kern-.025em b}\kern-.08em
    T\kern-.1667em\lower.7ex\hbox{E}\kern-.125emX}}
\begin{document}
\pagenumbering{arabic}
\title{Simultaneous face detection and 360 degree head pose estimation
\\
% \thanks{1: University of Engineering and Technology (VNU-UET) }
}

\makeatletter
\newcommand{\linebreakand}{%
  \end{@IEEEauthorhalign}
  \hfill\mbox{}\par
  \mbox{}\hfill\begin{@IEEEauthorhalign}
}
\makeatother

\author{\IEEEauthorblockN{Hoang Nguyen Viet}
\IEEEauthorblockA{UET AILab, VNU\\
Hanoi, Vietnam}
\IEEEcompsocitemizethanks{\IEEEcompsocthanksitem\IEEEauthorrefmark{2}corresponding author}
\and
\IEEEauthorblockN{Linh Nguyen Viet}
\IEEEauthorblockA{UET AILab, VNU\\
Hanoi, Vietnam}
\and 
\IEEEauthorblockN{Tuan Nguyen Dinh}
\IEEEauthorblockA{UET AILab, VNU\\
Hanoi, Vietnam}
\linebreakand
\IEEEauthorblockN{Duc Tran Minh}
\IEEEauthorblockA{UET AILab, VNU\\
Hanoi, Vietnam}
\and
\IEEEauthorblockN{Long Tran Quoc\IEEEauthorrefmark{2}}
\IEEEauthorblockA{UET SISLAB, VNU\\
Hanoi, Vietnam}}

\maketitle

\begin{abstract}

With many practical applications in human life, including manufacturing surveillance cameras, analyzing and processing customer behavior, many researchers are noticing face detection and head pose estimation on digital images. A large number of proposed deep learning models have state-of-the-art accuracy such as YOLO, SSD, MTCNN, solving the problem of face detection or HopeNet, FSA-Net, RankPose model used for head pose estimation problem. According to many state-of-the-art methods, the pipeline of this task consists of two parts, from face detection to head pose estimation. These two steps are completely independent and do not share information. This makes the model clear in setup but does not leverage most of the featured resources extracted in each model. In this paper, we proposed the Multitask-Net model with the motivation to leverage the features extracted from the face detection model, sharing them with the head pose estimation branch to improve accuracy. Also, with the variety of data, the Euler angle domain representing the face is large, our model can predict with results in the $360^{\circ}$ Euler angle domain. Applying the multitask learning method, the Multitask-Net model can simultaneously predict the position and direction of the human head. To increase the ability to predict the head direction of the model, we change the representation of the human face from the Euler angle to vectors of the Rotation matrix.

\end{abstract}


\section{Introduction}
Many solutions based on deep learning methods have recently been published to detect face and estimate head pose through digital images. Typically, some papers estimate head pose through the 2D or 3D facial landmarks detection \cite{3D_face_r}, \cite{3DLM_HP}, \cite{3D_Face_s} while other ones directly use the neural network model to determine the head pose as \cite{FSANet} \cite{RankPose}. A lot of results have been recognized by researchers and applied to reality to analyze the human gaze. However, there are some limitations to the above methods. Most methods have a pipeline with two main steps: (1) determine the position of the face(s) in the image, and (2) use machine learning algorithms or deep learning models to estimate the pose of the face(s) cropped from the results of step 1. If step 2 uses algorithms based on facial landmarks like \cite{3D_face_r}, \cite{3D_HP}, the results are sensitive to change of one or some of the facial landmarks that are determined incorrectly. As for deep learning models, many methods are robust and give impressive results, improving the weaknesses in machine learning methods.\par
Building deep learning model methods to handle the problem of identifying human face orientation is currently giving good results. Some typical results such as FSA-net \cite{FSANet}, RankPose \cite{RankPose}. However, in these methods, the angular domain to represent head pose is limited in $[-90^{\circ},90^{\circ}]$ or $[-99^{\circ}, 99^{\circ}]$ degree. One of the main reasons is that the neural network model built by these methods only learns to extract the features presented on the face. So when the face is rotated beyond the limit (here is $90^{\circ}$ or $99^{\circ}$), extracted features are not enough to make the prediction results no longer as accurate as usual.\par
Representing the face orientation in 3D space as three angles yaw, pitch, and roll is the most intuitive way on the image. These three angles correspond to the three primary ways in which the head can be rotated. This is commonly used in deep learning modeling problems for representing faces such as FSA-Net \cite{FSANet}, RankPose \cite{RankPose}. However, these angles have limitations. They create a drawback pointed out in article \cite{EA2RM} and it is called “gimbal lock.” Specifically, when two of the three axes representing pose are parallel, the other axis can not be determined. The function can map a face pose with too many angle representations. This causes mistakes during training model. The model after being trained will give unstable results and have a large variance.\par
To solve above problems, we propose a multitask learning method for the model to detect faces and estimate head poses simultaneously. In this paper, we use a dataset with a large Euler angle domain to represent head pose to train models with the hope that our model is objective.Then, we use basic vectors from rotation matrix to represent head pose instead of Euler angles to solve "gimbal lock" problem.\par 

\section{Related works}

\textbf{Face Detection}. 
Detecting face(s) on an image is a specific problem in the field of object detection. This problem aim is to develop models or algorithms to provide basic human-like observations to the computer, help the computer answer the question: ``Where is the object in the image ?". \cite{OD_survey} addresses the development history of object detection methods at deep learning. It shows that deep learning models are divided into two main groups: single-stage models and two-stages models. \cite{MimicDet} addresses differences in structure, pipeline, and the advantages - disadvantages of them. The single-stage model uses features extracted from the backbone to determine the location of the object and its class simultaneously. Some represent studies in this group are YOLOV3 \cite{yolov3}, SSD \cite{ssd}, Retina-Net \cite{focalloss}. Meanwhile, the two-stages model separates out the tasks and processes them independently. MTCNN \cite{mtcnn}, Faster R-CNN \cite{Frcnn} have the process to get started from feature extraction in the backbone, object detection, and finally object classification. Two-stages models generally give better accuracy but are slower about inference time due to the larger weight size when compared to single-stage models. Currently, single-stage models are gradually improving their rank on the accuracy chart while keeping the model size the same.\par

\textbf{Head Pose estimation}. \cite{3D_face_r}, \cite{3DLM_HP}, \cite{3D_Face_s}, \cite{3D_HP}, \cite{HP_LM} solve this problem by machine learning methods with head pose representation is Euler angles calculated from 2D facial landmarks or 3D facial landmarks detected in face. HopeNet \cite{Hopenet} uses two adaptive graph convolutional neural networks to detect 2D facial landmarks, convert them to 3D and predict Euler angles represent head pose. Later, methods using deep learning gradually replaced the old method because of the ability to extract features in the faces automatically and directly estimate the head pose angles through a linear regression approach. FSA-Net \cite{FSANet} proposes fine-grained structure mapping to group features and aggregate them before taking them into linear regression to predict head pose. Other hands, Rank-Pose[10] uses Siamese \cite{siamese} to set up input relationships to improve the accuracy of prediction.\par

\textbf{Multitask learning}. The paper \cite{MTL_survey} takes an overview survey about multitask learning, gives a definition that compares multitask learning with transfer learning, multi-label learning. It also proposes methods to design and optimize the model efficiently. At the same time, this paper also classifies models based on multitask learning according to the feature learning approach with two classification directions: backbone structure or last layers structure. According to the backbone, there are two main types: Hard parameter sharing and Soft parameter sharing. Another way following the last layers structure, they are divided into Encoder-focused models and Decoder-focused models. Deep learning models built on multitask learning such as \cite{MTL_deep} or cross-stitch Network \cite{MTL_cross} to leverage common features extracted from the backbone. In Head pose estimation, Multitask learning approach has been studied recently. \cite{HP_LM}, \cite{Hyperface} propose to multitask models to predict head pose and facial landmarks together. However, there is still no research applying multitask learning for face detection, and head pose estimation.\par

\textbf{Head pose representation}. The method of converting from Euler angles which represent head pose estimation (yaw, pitch, roll) to rotation matrix is analyzed in \cite{EA2RM} and the strength and weakness of each type, especially the “gimbal lock” phenomenon when using Euler angle to represent the head pose. Since then, \cite{Vector_base}, \cite{vector_deep} apply the results obtained from above to use vectors from rotation matrix and build loss function, optimization method, data preprocessing for the training process, model evaluation for head pose estimation task .\par

After researching, we found that the previous face-oriented estimation problems would need two models, one for face detection and one for head pose estimation. Therefore, the face's features in the image will have to be extracted twice through each model. It would be optimal if the features obtained by face detection were directly applied to the head pose estimation to reduce the computation in neural network but still ensure good output.\par

\begin{figure*}[ht]
    \centering
    \includegraphics[width=0.85\textwidth]{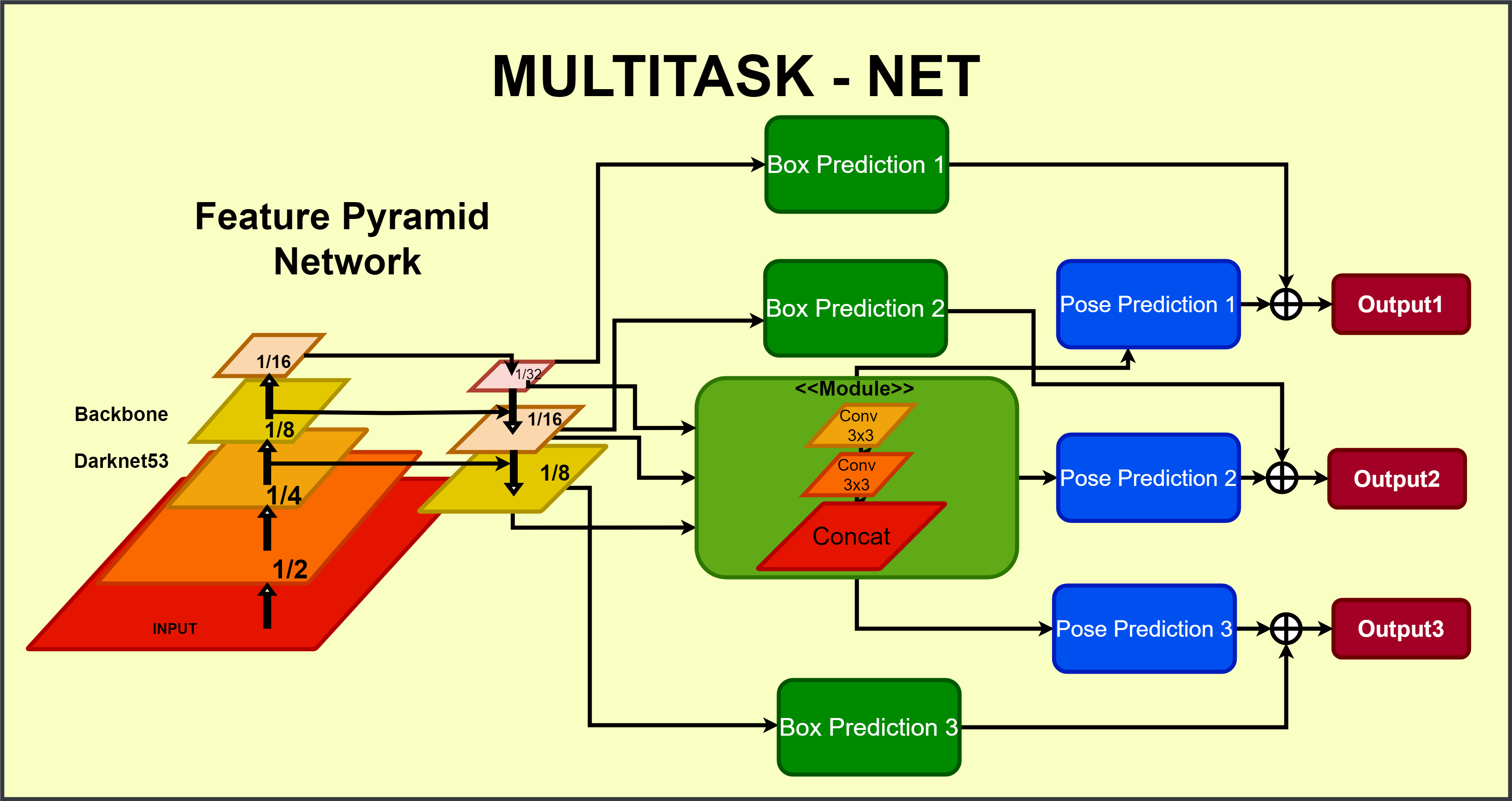}
    \captionsetup{justification=centering}
    \caption{The detailed architecture of Multitask-Net model.}
    \label{fig:systemoverall}
\end{figure*}

\section{Proposed Method}

% A.Xây dựng pipeline thu từng module 
In this section, we present our proposed model. We then introduce vectors to represent head pose and loss function applied to the vector during training.\par

\subsection*{3.1 Propose model}

We propose the Multitask-Net model which has a structure in Fig.\ref{fig:systemoverall}. This model is based on multitask learning with a backbone designed according to the Hard parameter sharing architecture, and the last layers following Decoder-focused architecture. With the idea inherited from YOLOv3 \cite{yolov3} about the face detection branch on the image, the backbone network of the model will extract features at different scales. This makes the model elicit many features of the object in the image with various sizes. When the features are successfully extracted, they are sent to the final convolution layers for decoding. To determine face position in the image, the features go through the face branch to calculate bounding boxes. From these predicted boxes, the model will tell us the position of the object on the image. We then aggregate the results of the face detection branch with features extraction from the backbone to calculate head poses corresponding to faces shown in the image. The aggregation module structure is present in Fig.\ref{fig:systemoverall}.\par

In the backbone of the Multitask-Net model, the Feature Pyramid Network (FPN) architecture \cite{fpn} is built with the Darknet-53 network as the kernel. The Darknet-53 \cite{yolov3} network is reported in paper \cite{yolov3} for remarkable results. It is made up of $53$ convolution layers with kernels size ($1$ x $1$) and ($3$ x $3$). As an advanced version of Darknet-19, the Darknet-53 network provides powerful features extraction. In many cases, this network is more efficient than today's commonly used backbones such as Resnet-101 or Resnet-152 \cite{Resnet}. \cite{yolov3} shows the results on different aspects to compare networks, and the Darknet-53 network gives the same accuracy as the Resnet-101 and Resnet-152 networks while using less GPU (due to the number of computations) and run faster. We use the Darknet-53 network in the FPN architecture with the function of feature extraction from coarse to meaningful.\par

After extracting features by the FPN network, these features will be passed through the last layers to decode and give prediction results. On the side of the face detection branch, we keep the same classes and operations according to \cite{yolov3}. After decoding features successfully, we have three feature maps corresponding to three different scales. Model passes them into the final convolution layers to bring the prediction results following format:\par 
$$Box Prediction_i = (bs, 3*(5+cls), K, K) \quad i\in\{1, 2, 3\}$$
% Grid unit = $1/13$ of input size: $(bs, 3x(5+cls), 13, 13)$\par
% Grid unit = $1/26$ of input size: $(bs, 3x(5+cls), 26, 26)$\par 
% Grid unit = $1/52$ of input size: $(bs, 3x(5+cls), 52, 52)$\par 
Where $K\in\{13, 26, 52\}$; $bs$ is the amount of input data, $cls$ is the number of object classifiers, $(K, K) = \{(13, 13), (26, 26), (52, 52)\}$ is the size of the image when dividing it into grid cells. For each grid cell, the image will be resized in grid cells (like in \cite{yolov3}) instead of pixels. The scalar $3$ corresponds to the number of anchor boxes. The scalar $5$ corresponds to the number of parameters presented object information, which are the coordinates of the center point, height, width, and confidence value of the model, indicating the probability of that box is positive. After extracting features and get features with 3 different ratios, we pass them through two parallel branches at the same time. In the first branch, the features will be rearranged and passed through the sigmoid function to get the information of positive boxes and negative boxes on the image (Box Predictions in Fig.\ref{fig:systemoverall}). This branch is called ``face detection branch". In the head pose estimation branch the features are moved to an aggregation module (green module in Fig.\ref{fig:systemoverall}). In the aggregation module, we pass them through two convolution and one concatenation layer to increase meaningful information and get better accuracy (the results of this branch are Pose Predictions in Fig.\ref{fig:systemoverall}). We concatenate the results of the two branches once they have finished running (outputs in Fig.\ref{fig:systemoverall}) and get the results with the following format:\par

$$Output_i = (bs, 3*(5+cls+np), K, K) \quad i\in\{1, 2, 3\} $$

% Grid unit = $1/13$ of input size: $(bs, 3x(5+cls+np), 13, 13)$ \par
% Grid unit = $1/26$ of input size: $(bs, 3x(5+cls+np), 26, 26)$ \par
% Grid unit = $1/52$ of input size: $(bs, 3x(5+cls+np), 52, 52)$ \par

Where $np$ denotes the number of parameters which present head pose.\par

\subsection*{3.2 Rotation matrix}

\textbf{Convert yaw, pitch, roll to rotation matrix}

We have referred to the conversion as in \cite{Vector_base}. This paper uses three rotation matrices with kernels corresponding to three angles: yaw, pitch, roll, multiply them to get a matrix with three columns: three orthogonal vectors representing three dimensions: x-axis, y-axis, and z-axis. Suppose the matrix $R = [r_1, r_2, r_3]^T$ where $r_i$ is the $i^{th}$ column vector of the $R$ matrix. And we define the three left, bottom and front vectors of the point of view as $v_1 = [1, 0 , 0]^T$, $v_2 = [0, 1, 0]^T$, $v_3 = [0, 0, 1]^T$. From there, three representation vectors will be: $v'_1 = Rv_1 = r_1$, $v'_2 = Rv_2 = r_2$, $v'_3 = Rv_3 = r_3$. There are $12$ different ways to convert three angles yaw, pitch, and roll to vectors from rotation matrices. Since they are equivalent, we choose a matrix to solve the head pose estimation task. Specifically, the three rotation matrices would be:\par

$$R_z = \begin{bmatrix}
            1 & 0 & 0 \\
            0 & \cos{roll} & -\sin{roll} \\
            0 & \sin{roll} & \cos{roll} \\
\end{bmatrix}$$

$$R_y = \begin{bmatrix}
            \cos{pitch} & 0 & \sin{pitch} \\
            0 & 1 & 0 \\
            -\sin{pitch} & 0 & \cos{pitch} \\
\end{bmatrix}$$

$$R_x = \begin{bmatrix}
            \cos{yaw} & -\sin{yaw} & 0 \\
            \sin{yaw} & \cos{yaw} & 0 \\
            0 & 0 & 1 \\
\end{bmatrix}$$ 

From the rotation matrix and normalizing the elements in the matrix, we find 3 unit vectors corresponding to x-axis, y-axis, and z-axis. these vectors make an orthogonal basis.\par

\textbf{Convert rotation matrix to  yaw, pitch, roll }
 	
As mentioned, Any double in three vectors are orthogonal. So when using the vector representation method for determining the initial direction, we add a loss function to constrain the orthogonality of these three vectors. Despite having the loss function, the Rotation Matrix R predicted by the model is still uncertain to preserve the properties of these vectors. Therefore, according to the paper \cite{Vector_base}, it is necessary to find the Rotation Matrix containing three orthogonal unit vectors  "closest" to the three predicted vectors for the validation process. Here,``closest" is defined as the sum of the shortest euclidean distances between pairs of vectors. To search for the matrix $R$ with the above condition, we use SVD (singular value decomposition) to separate matrix $R$ = $UEV^T$ where $U$ and $V$ are orthogonal matrices, $E$ is the diagonal matrix. The matrix $R'$ satisfied will be $R'$ = $UV^T$. In this way, $det(R')$ can be equal to -1, so the final formula that defines the matrix $R'$ is:\par
 	$$R' = U*diag(1,1,-1)*V^T$$ 
With $diag(1,1,-1)$ is a diagonal matrix with the elements on the diagonal are 1, 1, -1, respectively.\par

\subsection*{3.3 Multitask loss}

As written in section 3.1, the return result of the branch determining the face position is 4 coefficients representing the bounding box, confidence coefficient and class distribution. Each coefficient (or pair of coefficients) will have its own loss function, so this branch has a total of 4 loss functions: \par
\begin{equation}
Loss_{bbox} = \lambda_{xy}L_{xy} + \lambda_{wh}L_{wh} + \lambda_{cls}L_{cls} + \lambda_{obj}L_{obj}
\end{equation}

Where  $\lambda_{xy}$, $\lambda_{wh}$, $\lambda_{cls}$, $\lambda_{obj}$ represent the ratio of the specific loss function of each part of the total loss function. \par

When using 3 vectors made up of rotation matrices to represent the face orientation, the loss function used for this result will also have a different structure when using 3 angles to represent the face. We use the MSE loss function for the Euclidean distance between the two vectors. Specifically, the function structure would be follow:  \par
\begin{equation}
L_{vmse}(v_{pred}, v_{true}) = \sum_{i=0}^{dim} (v^{i}_{pred} - v^{i}_{true})^{2}
\end{equation}

where dim is the dimensionality of the vector. Since the special condition of these three vectors is that all of pairs from them are orthogonal , the paper adds a loss function to force this condition between the vectors: \par 
\begin{equation}
L_{ortho} = \sum_{i != j}^{dim}L_{mse}(v^{i}v^{j}, 0)
\end{equation}

Then the loss function for head pose estimation branch is: \par 
\begin{equation}
\begin{split}
L_{pose} = L_{mse}(v_{x\_pred}, v_{x\_true})  + \\
L_{mse}(v_{y\_pred}, v_{y\_true}) + \\
L_{mse}(v_{z\_pred}, v_{z\_true}) + L_{ortho}
\end{split}
\end{equation}

Essentially, as many other multitask learning studies have used, our aggregate loss function is made up of the sum of the losses of all branches with proportions corresponding to the importance of that branch. in models: \par
\begin{equation}
Loss = \alpha L_{bbox} + (1 - \alpha)L_{pose}
\end{equation}

where $\alpha$ denotes loss ratio of a branch in total loss. \par

\section{Experiment}
In this section, the experiments will show the results and the evaluation of the proposed models. Before going into the detailed evaluation of the models, we will outline how to conduct the experiment and the evaluation methods. Specifically, Section 4.1 shows the training strategy. Section 4.2 describes the data and the evaluation protocols. Section 4.3 shows the result of the model evaluation mentioned in section 3.
\subsection*{4.1 Training strategy}
The training process is divided into three phases:\par
\begin{itemize}
    \item First phase (first $50$ epochs, batch size = $64$, learning rate = $1e-3$): we freeze the backbone and head pose estimation branch, training only the face detection last layers.
    \item Second phase ($50$ next epochs, batch size = $16$, learning rate = $1e-4$): we continue training the face detection branch and start training the backbone; still free head pose estimation branch. The purpose of stages 1 and 2 is to increase the ability to extract features and detect faces more exactly in the image before estimating the head poses.
    \item Third phase (last $50$ epochs, batch size = $16$, learning rate = $1e-4$): the model will be fully trained from backbone to feature branches without freezing any layer of the model.
\end{itemize}

\subsection*{4.2 Dataset and evaluation protocols}

\begin{table}[t]
    \captionsetup{type=figure, position=above, justification=centering}
    \captionof{table}{Evaluation models which are trained in 300WLP datasets in the all BIWI dataset} \label{tab:evaluate_BIWI}
\begin{center}
\begin{tabular}{lcccc}
\hline
\multicolumn{1}{|l|}{\textbf{Model}}            & \multicolumn{1}{l|}{\textbf{Yaw}}  & \multicolumn{1}{l|}{\textbf{Pitch}} & \multicolumn{1}{l|}{\textbf{Roll}} & \multicolumn{1}{l|}{\textbf{MAE}} \\ \hline

\multicolumn{1}{|l|}{KEPLER\cite{KEPLER}} & \multicolumn{1}{c|}{8.8}           & \multicolumn{1}{c|}{17.3}           & \multicolumn{1}{c|}{16.2}          & \multicolumn{1}{c|}{13.9}          \\ \hline

\multicolumn{1}{|l|}{Dlib\cite{dlib}} & \multicolumn{1}{c|}{16.8}           & \multicolumn{1}{c|}{13.8}           & \multicolumn{1}{c|}{6.19}          & \multicolumn{1}{c|}{12.2}          \\ \hline

\multicolumn{1}{|l|}{FAN\cite{FAN}} & \multicolumn{1}{c|}{8.53}           & \multicolumn{1}{c|}{7.48}           & \multicolumn{1}{c|}{7.63}          & \multicolumn{1}{c|}{7.89}          \\ \hline

\multicolumn{1}{|l|}{Hopenet(a=2)\cite{Hopenet}} & \multicolumn{1}{c|}{5.17}           & \multicolumn{1}{c|}{6.98}           & \multicolumn{1}{c|}{3.39}          & \multicolumn{1}{c|}{5.18}          \\ \hline

\multicolumn{1}{|l|}{Hopenet(a=1)\cite{Hopenet}} & \multicolumn{1}{c|}{4.81}           & \multicolumn{1}{c|}{6.61}           & \multicolumn{1}{c|}{3.27}          & \multicolumn{1}{c|}{4.9}          \\ \hline

\multicolumn{1}{|l|}{SSR-Net-MD\cite{SSR}} & \multicolumn{1}{c|}{4.49}           & \multicolumn{1}{c|}{6.31}           & \multicolumn{1}{c|}{3.61}          & \multicolumn{1}{c|}{4.65}          \\ \hline

\multicolumn{1}{|l|}{FSA-Net\cite{FSANet} (Fusion)} & \multicolumn{1}{c|}{4.27}           & \multicolumn{1}{c|}{4.96}           & \multicolumn{1}{c|}{2.76}          & \multicolumn{1}{c|}{4.00}          \\ \hline

\multicolumn{1}{|l|}{WHENet \cite{WHENet}}         & \multicolumn{1}{c|}{5.11}          & \multicolumn{1}{c|}{6.24}           & \multicolumn{1}{c|}{4.92}          & \multicolumn{1}{c|}{5.42}          \\ \hline

\multicolumn{1}{|l|}{Multitask-Netv2(euler angle)}     & \multicolumn{1}{c|}{4.64}          & \multicolumn{1}{c|}{7.23}           & \multicolumn{1}{c|}{6.23} & \multicolumn{1}{c|}{6.03}        \\ \hline

\multicolumn{1}{|l|}{\textbf{Multitask-Netv2(vector base)}}     & \multicolumn{1}{c|}{\textbf{4.62}}          & \multicolumn{1}{c|}{\textbf{4.29}}           & \multicolumn{1}{c|}{\textbf{4.52}} & \multicolumn{1}{c|}{\textbf{4.48}}          \\ \hline

\end{tabular}   
% \label{tab:evaluate_BIWI}
\end{center}
\end{table}

Training dataset: We use CMU dataset \cite{CMU} with $81$ sequences; each sequence includes $31$ videos. In each video, there are one or more people in single frame annotated. After processing we get close to $400k$ images, all of them are annotated and have a range of yaw angle = $[-179^{\circ}, 179^{\circ}]$, pitch = $[-90^{\circ}, 90^{\circ}]$, roll = $[-90^{\circ}, 90^{\circ}]$. In addition, we use a set of 300WLP \cite{300WLP} and a BIWI \cite{BIWI} part with yaw, pitch, and roll angular domains of $75^{\circ}$, $60^{\circ}$, and $50^{\circ}$, respectively, which only is a single person image annotated to compare with other methods. \par

Testing dataset: the dataset mainly used for testing is BIWI datasets. In addition, we divide a part of the video from sets of CMU datasets to evaluate the model more objectively because so far, there are only CMU datasets with multiple people in single image format be annotated for face detection \& head pose estimation.\par
Evaluation protocol: we follow the protocol of FSA-Net \cite{FSANet} - The model uses CNN for head pose estimation compared with many other models currently available. FSA-Net proposes two protocols: (1) train on 300 WLP test on BIWI, (2) trains on $70\%$ BIWI, test on $30\%$ BIWI. In addition, we train on the CMU dataset to compare the face-oriented representations. We use IoU metric for head detection and MAE loss for head pose estimation in all our tables\par

\begin{figure*}[ht]
    \centering
    \includegraphics[width=1\textwidth]{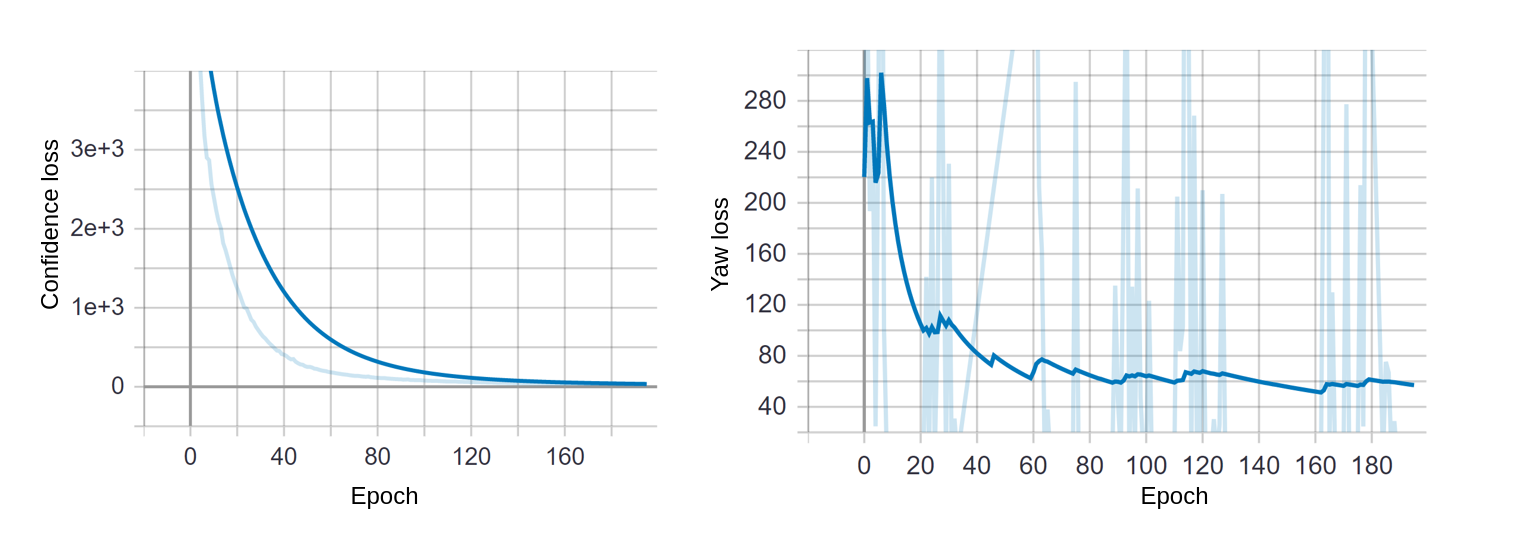}
    \captionsetup{type=figure, position=above, justification=justified}
    \caption{Loss graphs of Multitask-Net model when training all branch. In this figure, confidence loss graph of face detection branch is on the left and Yaw loss of head pose estimation branch is on the right.}
    \label{fig:loss}
\end{figure*}

\subsection*{4.3 Results}

After training according to the strategy in session 4.1 with $55$ minutes per epoch with CMU dataset or 300 WLP ($15$ minutes per epoch) for both face detection and head pose estimation branch, the loss function results on both branches gradually decrease over each epoch and they are shown in  Fig.\ref{fig:loss}. After training through all the phases, The Confidence loss of face detection (on the left of Fig.\ref{fig:loss}) and the Yaw loss of head pose (on the right of Fig.\ref{fig:loss}) decrease with each epoch and converge at the 180th epoch.\par

We compare our method with those that stand out today. The first group is to estimate the face orientation from the position of the facial landmarks. KEPLER \cite{KEPLER} predicts the orientation and position of facial landmarks at the same time as the GoogleLeNet architecture \cite{googlenet} added some transformations. FAN \cite{FAN}, Dlib \cite{dlib} were also very popular in this group because of their fast prediction speed. The second is that groups directly use features extracted from deep learning neural networks for head pose estimation. Representative example is Hopenet \cite{Hopenet} with ResNet backbone, MSE loss function combined with cross-entropy to train the model according to the regression problem. Or Fine-Grained Structure Aggregation at different levels for SSR-Net. Especially FSA-Net – the model proposed in 2019 with the improvement from SSR-Net and the addition of the capsule module helps the model predict high accuracy results. Besides, RankPose with a different direction from the above models also gave positive results. This model applies Siamese architecture to learn more relationships between images to improve features extracted from the model. We train the model with a variety of data types to compare multiple existing methods.

In Table \ref{tab:evaluate_BIWI}, our model gives optimistic results, overcome all already methods in the group using the facial landmarks 2D or 3D points as the basis for estimation of face orientation. Although the model proposed in the paper is not equal to some robust models such as FSA-Net, VGG16, the partial reason is an error of the conversion from the rotation matrix to the three Euler angles. If the models are applied with the representation of the vectors of the rotation matrix, the model we propose is no less than the models mentioned above.
\begin{table}[ht]
    \captionsetup{type=figure, position=above, justification=centering}
    \captionof{table}{Compare our different version models which are trained in 70\%BIWI dataset in the 30\% BIWI dataset} \label{tab:compare_own_model}
\begin{center}
\begin{tabular}{lcccc}
\hline
\multicolumn{1}{|l|}{\textbf{Model}} & \multicolumn{1}{l|}{\textbf{IoU}} & \multicolumn{1}{l|}{\textbf{Yaw}}  & \multicolumn{1}{l|}{\textbf{Pitch}} & \multicolumn{1}{l|}{\textbf{Roll}}      \\   \hline

\multicolumn{1}{|l|}{Multitask-Netv1} & \multicolumn{1}{c|}{63.8}  & \multicolumn{1}{c|}{4.53}  & \multicolumn{1}{c|}{4.6} & \multicolumn{1}{c|}{3.47}    \\     \hline

\multicolumn{1}{|l|}{Multitask-Netv1 (tanh function)} & \multicolumn{1}{c|}{73.4}  & \multicolumn{1}{c|}{5.49}  & \multicolumn{1}{c|}{3.92} & \multicolumn{1}{c|}{3.21}      \\     \hline

\multicolumn{1}{|l|}{Multitask-Netv2 (euler angle)} & \multicolumn{1}{c|}{60.72}   & \multicolumn{1}{c|}{6.02}   & \multicolumn{1}{c|}{5.33} & \multicolumn{1}{c|}{5.11}     \\       \hline

\multicolumn{1}{|l|}{\textbf{Multitask-Netv2 (vector base)}} & \multicolumn{1}{c|}{\textbf{63.6}}   & \multicolumn{1}{c|}{\textbf{5.33}} & \multicolumn{1}{c|}{\textbf{3.9}}    &       \multicolumn{1}{c|}{\textbf{3.28}}    \\     \hline

\end{tabular}   
% \label{tab:compare_own_model}
\end{center}
\end{table}

As shown in Table \ref{tab:compare_own_model}, we train different version models and evaluate them in BIWI dataset, with Multitask-Netv1 (Multitask-Net version 1) being simple multitask model and its two branches are designed not separate. Then we normalize output of face detection branch by tanh function to project arbitrary range to $[-1, 1]$. the IoU result of model evaluation after using tanh function is $73.4$ better than old model ($63.8$). In the end, we build model with tanh function for output of face detection branch and represent head pose with based vectors from rotation matix for head pose estimation branch, all branches are designed deeply and dependently. This powerful model called \textit{Multitask-Netv2} gives predictions expressively and is shown in last line of Table \ref{tab:compare_own_model}.

Both the BIWI and 300WLP datasets contain mostly portraits and only one person per image, and the person-to-image distance does not change (in the BIWI dataset). This makes it difficult to evaluate the proposed model using only these two datasets objectively. Therefore, we use the CMU to get another perspective on the proposed model. This data set has many people in the image, and the structure is a meaningful scenery that helps the model to evaluate both face detection and head pose estimation. Table \ref{tab:cmu_compare} shows the results of the proposed models of both branches on the CMU dataset. Models with positive results when using the yaw angle domain are $[-179^{\circ}, 179^{\circ}]$. The multitask-netV2 model with the head pose representation as the vector of the rotation matrix gives better results than the one using the three angles yaw, pitch, roll. As there is currently no model that uses the CMU dataset to predict multiple people in an image simultaneously as the proposed method, we do not compare our model with other models on this dataset.

\begin{table}[ht]
    \captionsetup{type=figure, position=above, justification=centering}
    \captionof{table}{Evaluate our models trained in CMU dataset} \label{tab:cmu_compare}
\begin{center}
\begin{tabular}{lcccc}
\hline
\multicolumn{1}{|l|}{\textbf{Model}} & \multicolumn{1}{l|}{\textbf{IoU}} & \multicolumn{1}{l|}{\textbf{Yaw}}  & \multicolumn{1}{l|}{\textbf{Pitch}} & \multicolumn{1}{l|}{\textbf{Roll}}      \\   \hline

\multicolumn{1}{|l|}{Multitask-Netv2 (euler angle)} & \multicolumn{1}{c|}{59.07}   & \multicolumn{1}{c|}{11.49}   & \multicolumn{1}{c|}{13.26} & \multicolumn{1}{c|}{9.45}     \\       \hline

\multicolumn{1}{|l|}{\textbf{Multitask-Netv2 (vector base)}} & \multicolumn{1}{c|}{\textbf{62.31}}   & \multicolumn{1}{c|}{\textbf{9.55}} & \multicolumn{1}{c|}{\textbf{11.29}}    &       \multicolumn{1}{c|}{\textbf{8.32}}    \\     \hline

\end{tabular}   
% \label{tab:cmu_compare}
\end{center}
\end{table}

\section{CONCLUSION}
Paper presented ideas and methods related to the field of face detection and head pose estimation. In addition, a multitask learning model is proposed for the problem of simultaneously determining the position and direction of the head. The proposed model has achieved high accuracy in both of 2 problems. The results are comparable with other popular state-of-the-art methods and, at the same time, give two results to increase the applicability of the problem. Especially with the module for building fast problems in the direction of multitask learning that can be combined, used with many different feature extraction models with positive results.\par
The lack of diversity in the data perspective makes the objectivity of the model only within a certain range. Although we have used many different image enhancement methods to limit the feature similarity between images, making the model more objective, but not solving the whole problem. Therefore, the study of the problem of simultaneously determining the position and direction of the head needs to be further researched in many different directions, such as data diversity, flexibility in architecture. At the same time, the optimization of the model, avoiding waste of resources, should be paid attention to.
\section*{Acknowledgment}
This work has been supported by VNU University of Engineering and Technology. 

% \vspace{12pt}
% \color{red}
% IEEE conference templates contain guidance text for composing and formatting conference papers. Please ensure that all template text is removed from your conference paper prior to submission to the conference. Failure to remove the template text from your paper may result in your paper not being published.

\end{document}